\documentclass{article}

\usepackage[final]{nips_2017}

\usepackage[utf8]{inputenc} 
\usepackage[T1]{fontenc}    
\usepackage{hyperref}       
\usepackage{url}            
\usepackage{booktabs}       
\usepackage{amsfonts}       
\usepackage{nicefrac}       
\usepackage{algorithm}
\usepackage[noend]{algpseudocode}
\usepackage{float}
\usepackage{amsmath}
\usepackage{amssymb}

\usepackage{graphicx,pgfplots,verbatim,adjustbox,amsmath}            \usepackage{mathtools}
\usepackage{algorithm}
\usepackage[noend]{algpseudocode}

\algrenewcommand\algorithmicforall{\textbf{foreach}}
\algrenewcommand\algorithmicindent{.8em}
\usepackage{array}
\usepackage{eqparbox}
\usepackage{array}
\usepackage{eqparbox}

\usepackage{microtype}      
\makeatletter
\def\BState{\State\hskip-\ALG@thistlm}
\makeatother
\makeatletter
\usepackage[noend]{algpseudocode}
\usepackage{float}
\algnewcommand\INPUT{\item[\textbf{Input:}]}%
\algnewcommand\OUTPUT{\item[\textbf{Output:}]}

\author{
  Housam Khalifa Bashier Babiker and Randy Goebel\\
Alberta Machine Intelligence Institute\\
  Department of Computing Science University of Alberta\\
  Edmonton, Alberta Canada T6G 2E8 \\
  \texttt{khalifab@ualberta.ca,rgoebel@ualberta.ca} \\
}

\title{Using KL-divergence to focus Deep Visual Explanation}

\begin{document}

\maketitle

\begin{abstract}
We present a method for explaining the image classification predictions of deep convolution neural networks, by highlighting the pixels in the image which influence the final class prediction. Our method requires the identification of a heuristic method to select parameters hypothesized to be most relevant in this prediction, and here we use Kullback-Leibler divergence to provide this focus. Overall, our approach helps in understanding and interpreting deep network predictions and we hope contributes to a foundation for such understanding of deep learning networks. In this brief paper, our experiments evaluate the performance of two popular networks in this context of interpretability.
 
\end{abstract}

\section{Introduction}

Deep convolution neural networks have shown impressive results in many domains such as in computer vision problems. The fundamental improvement of current deep learning methods is that, unlike earlier more shallow network layers, deep learning automatically identifies appropriate stratification of relevant attributes to construct a predictive model [2]. But despite improvements on quality and accuracy of predictions, these models are hard to interpret, especially when they become deeper and training weight adjustments propagate to fully connected layers. The problem of interpreting (or debugging) becomes extremely important in more sensitive domain such as dealing with medical data, i.e. any wrong decision might directly  affect patient's well-being. We therefore need to be able to build models which are interpretable and can support the explanation of their decisions. While neural networks can achieve high accuracy, they are not easily interpretable. Improved interpret-ability will help inform tradeoffs, e.g., methods like Bayesian nets are more easily interpretable but less accurate.

Recently, many methods have been proposed to address this trade-off  of interpret-ability [8-10]. One approach is based on computing the gradient of the output layer with respect to the feature maps [3], which is a relaxation generalization of [5]. The gradient is then back-propagated to a feature map layer, to try and distinguish the discriminative pixels for any input image. The gradient is determined based on network activation; to focus attention to where network activation provides insight into classifier output, we here consider two popular methods.

In the following, we explain our approach by describing a simple method to interpret deep neural network prediction. Our approach produces a map which highlights the salient region for the input image that represents the evidence related to the activations of the output layer of the network, i.e. why this network arrived at this decision?

The reminder of this paper is organized as follows. Section 2 presents our proposed approach. In Section 3 we describe the experiments with our initial framework, and  finally, Section 4 concludes with our preliminary results, followed by a discussion of extensive future work.

\section{Approach}
In the case of image classification, we focus on the loss of spatial information in fully connected layers of deep convolutional networks, and identify an approximate visual explanation using the feature maps $X$ at a particular convolution layer $l$. To be able to explain a prediction of the black box function $f$, we rely on computing the Kullback-Leibler (KL) divergence gradient of the raw class scores $y^{'}$ with respect to ground truth $y$, to estimate the discriminative localization map. We use this gradient as evidence for the network prediction. To achieve this, we  first compute the joint probabilities of  $y^{'}$ as follows:

\begin{equation}
\begin{split}
 p_{ij} =   \frac{(1+ ||k_{i}-k_{j}||^2)^{-1} }{\sum_{u \neq v} ( 1+||k_{u}-k_{v}||^2)^{-1}}
 \end{split}
\end{equation}

Here $p_{ij}$ denotes the joint probabilities, $k$ is the raw class scores before softmax , $i$ indexes a neuron value and  $\sum_{u \neq v}$ combines all the values. For the ground truth we estimate the pairwise affinities with perplexity.  We then compute the KL-divergence gradient i.e. $\frac{\delta y^{'}}{\delta y} \Rightarrow z$ derived here [6]. We also normalize the gradient to a zero mean and unit variance as follows:

\begin{equation}
\begin{split}
 \alpha =   \frac{z - \mu }{\sigma z} 
 \end{split}
\end{equation}

The obtained weights $\alpha$ capture the relevant information in the feature maps acquired by the network. These weights are applied to every feature map $x_{i} \in X$ as to identify the discriminative pixels which influence the final prediction output as follows:

\begin{equation}
\begin{split}
 E_{KL-divergence} =  \sum_{i}\sum_{j} x_{i} * |\alpha_{j}|
 \end{split}
\end{equation}

\begin{figure}[H]

\centering
\includegraphics[width=13cm,height=5.5cm]
{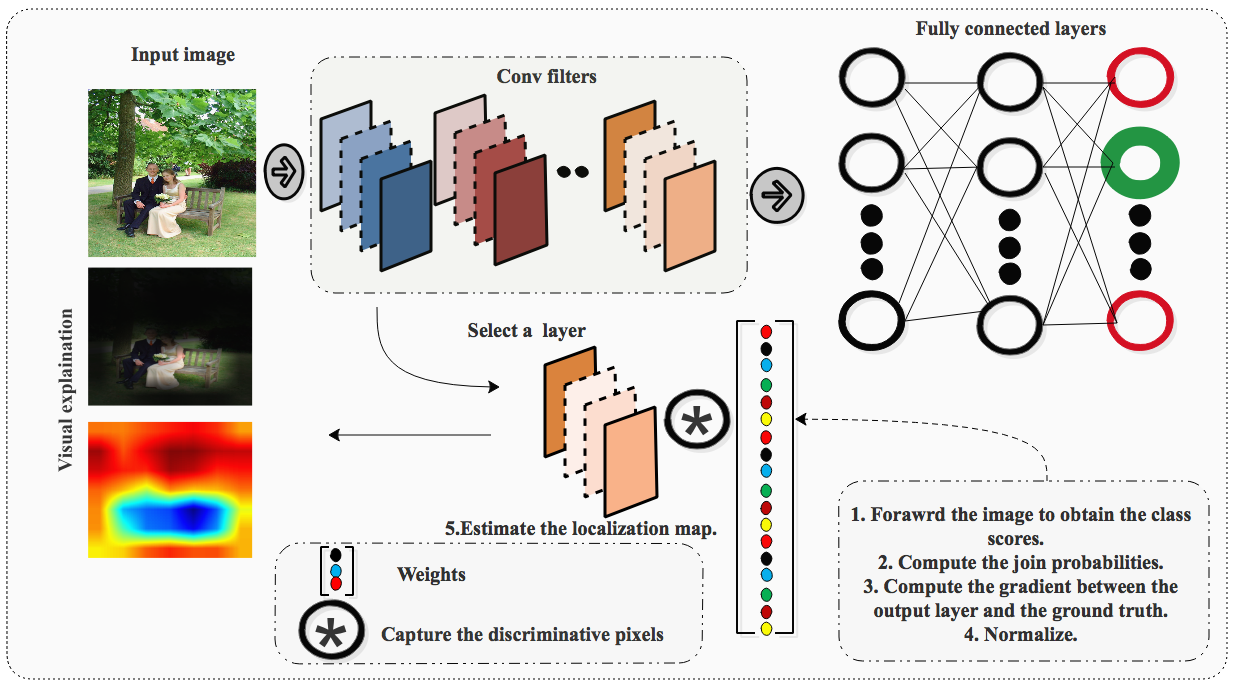}
\caption{Our proposed methodology.}
\label{fig:main_d}
\end{figure}

The $\sum_{i}$ runs over all the features $x_{i} \in X$ and $j$ runs over all the obtained weights $\alpha$. The output of (3) is the localization map which can be used as the evidence to explain the prediction of the black box function. Finally, we normalize the heat-map $E_{KL-divergence}$ for visualization. Algorithm 1 summarizes the implementation method and the overall methodology is depicted in Figure \ref{fig:main_d}.

\begin{algorithm}
\caption{Proposed approach}\label{algorithm1}
\begin{algorithmic}
\INPUT image, ground truth $y$
\OUTPUT Discriminative localization map  $\Rightarrow E_{KL-divergence}$\\
Apply a single forward-pass to estimate $\Rightarrow y^{'}$\\
Compute the joint probabilities for both $y^{'}$ and $y$\\
Compute the gradient and normalize using {\color{red} {$(2)$}} $\Rightarrow \alpha$ \\
initialize  $E_{KL-divergence}$ to zero
\For{$i = 1$ \textbf{to} ${nFeatureMaps}$}
\State Initialize temp to zero
\For{$j = 1$ \textbf{to} ${sizeof\alpha}$}
\State $temp \gets  temp +(x_{i}*|\alpha_{j}|)$  
\EndFor
\BState {\textbf{end for}}
\State $E_{KL-divergence} \gets E_{KL-divergence} + temp$
\EndFor

\BState {\textbf{end for}}

\end{algorithmic}
\end{algorithm}

\section{Experiments}
\begin{figure}[H]
\centering
\includegraphics[width=13cm,height=5cm]
{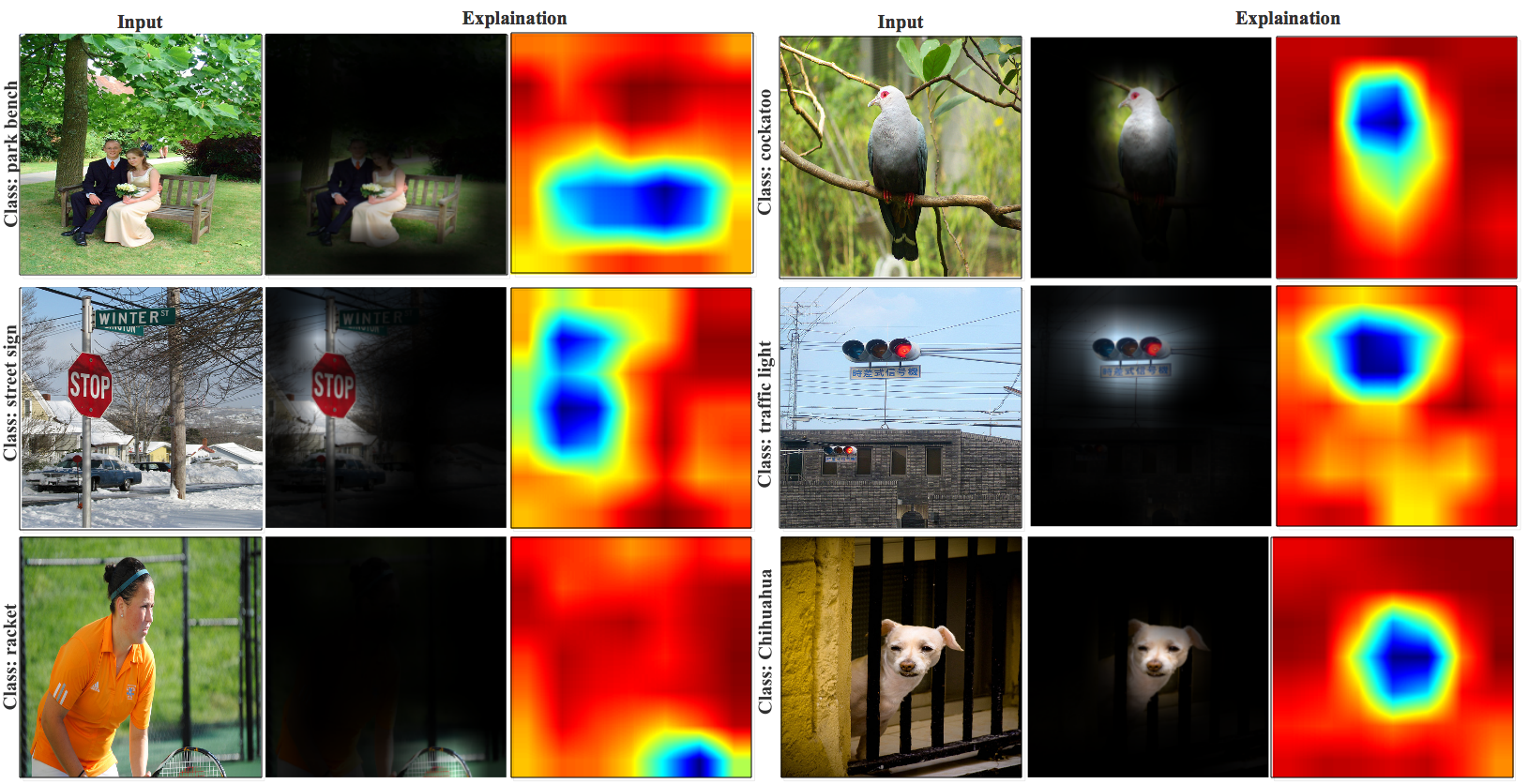}
\caption{Explaining the decisions made by the VGG-16 (park bench, street sign, racket, cockatoo, traffic light and chihuahua), our approach highlights the most discriminative region in the image.}
\label{fig:vggexamp}
\end{figure}

To evaluate the proposed method in the context of visual explanation, we select images from the common objects in context (COCO) challenge set [1]. For the learned models, we employed the publicly available pre-trained model VGG-16 [4] . The initial results of our visualization algorithm are illustrated in Figure \ref{fig:vggexamp}. \par
\begin{figure}[H]

\centering
\includegraphics[width=13cm,height=5cm]
{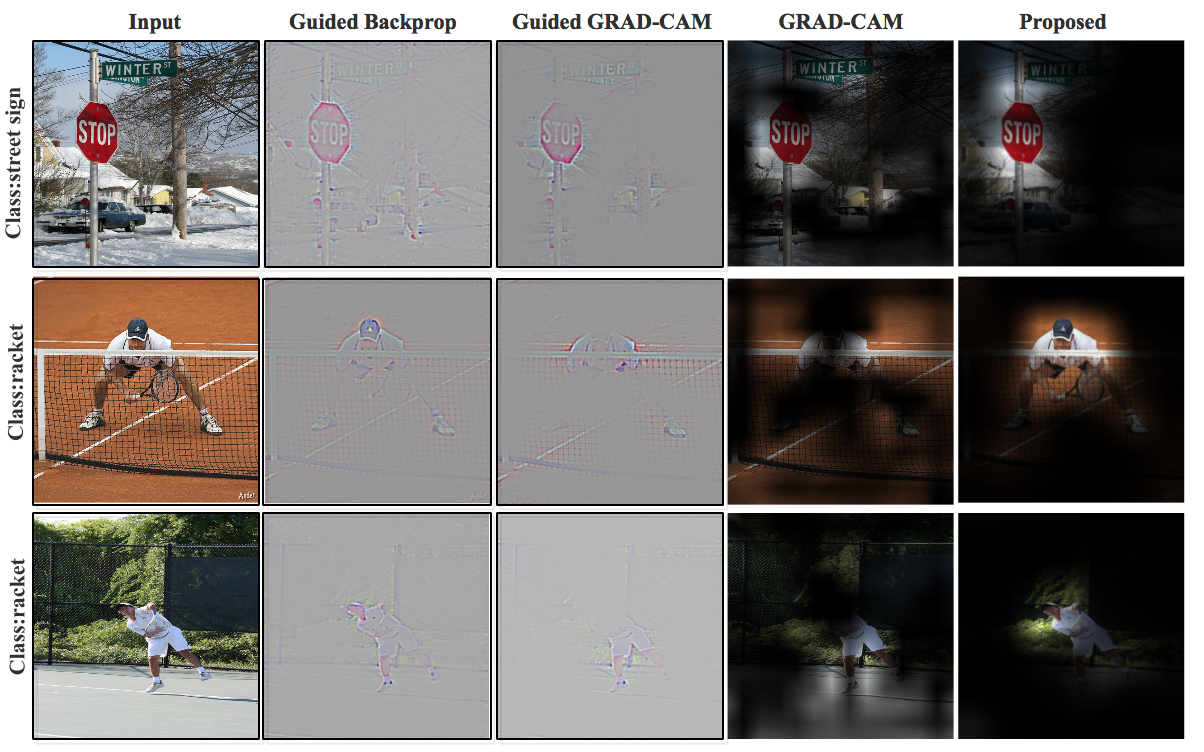}
\caption{Results on randomly chosen images from COCO visualizing VGG-16 predictions (street sign, racket and racket).}
\label{fig:compmethods}
\end{figure}

We also compared our explanation algorithm with other similar methods i.e., guided back- propagation [9] and gradient weighted class activation [3]. The comparison results are depicted in Figure \ref{fig:compmethods}.
\subsection{Comparing two deep neural networks}
\begin{figure}[H]

\centering
\includegraphics[width=13cm,height=5cm]
{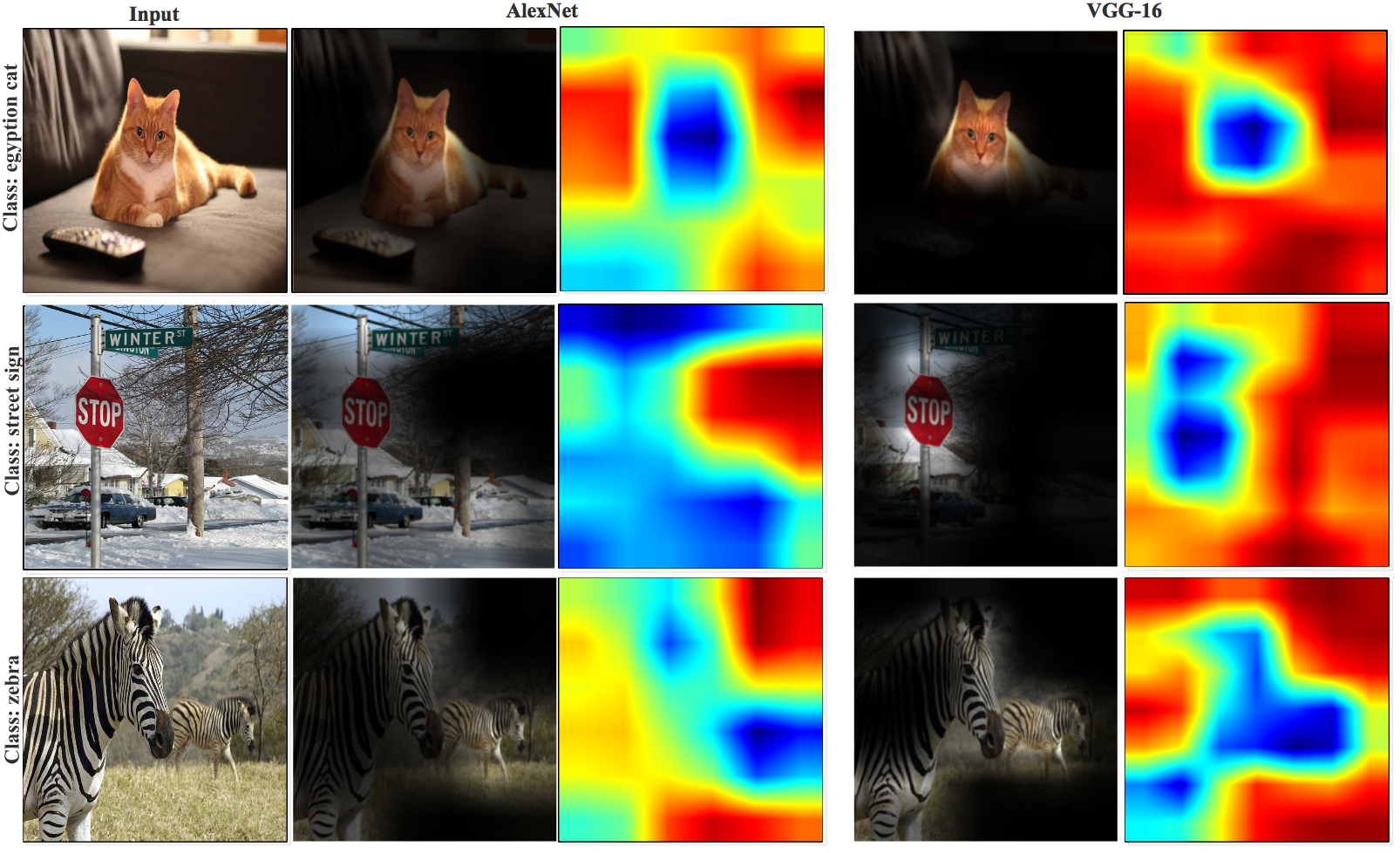}
\caption{Comparing visualization results of two networks.}
\label{fig:compnetworks}
\end{figure}

It is also important to understand the influence of the network architectures on the visualization algorithms. For the purpose of this experiment, we tested our algorithm on VGG-16 and AlexNet [7] networks. The comparison results are shown in Figure \ref{fig:compnetworks}. From these results, we can see that the quality of the visualization depends on the overall accuracy of the network. Since VGG-16 is more reliable, we can observe that the network pays attention to a smaller salient area, compared to the AlexNet network. Therefore, the better the accuracy the better the visualization interpretation.
\section{Conclusion}
We introduce a new approach to explaining deep neural networks prediction. This particular solution focuses identification of salient regions by computing the KL-divergence gradient from the ground truth label and the output layer. Our preliminary experiments on image classification problems compare the performance of two networks. We believe that this work can further be extended to evaluate decisions in a variety of applications of deep neural networks, in image classification, speech, and text.  We are especially keen to consider text domains, as interpretability will expose patterns of semantic relatedness beyond that of images.

\setcounter{secnumdepth}{0} 
\section{Acknowledgment} 

We thank colleagues from the Alberta Machine Intelligence Institute for their discussions and advice. This work was supported by NSERC and AMII.



\section*{References}
\medskip

\small
[1] Lin T.Y., Maire M., Belongie S., Hays J. Perona, P. Ramanan D., Dollár P. and Zitnick C.L. Microsoft coco: Common objects in context. In \textit{ECCV}, 2014.

[2] LeCun Y., Bengio Y., and Hinton G. Deep learning. \textit{Nature} pages 436-444, 2015.

[3] Selvaraju R. R., Das A., Vedantam R., Cogswell M. Parikh, D. and Batra D. Grad-CAM: Why did you say that?. In \textit{Workshop on Interpretable Machine Learning in Complex Systems, NIPS}, 2016.

[4] Simonyan K., and Zisserman A. Very deep convolutional networks for large-scale image recognition. \textit{arXiv preprint arXiv:1409.1556}, 2014.

[5] Zhou B., Khosla A., Lapedriza, A. Oliva, A. and Torralba A. (2016). Learning deep features for discriminative localization. In \textit{Proc. CVPR}, pages 2921-2929, 2016.

[6] Maaten L. V. D., and Hinton G. Visualizing data using t-SNE. \textit{Journal of Machine Learning Research}, pages 2579-2605, 2008.

[7]Krizhevsky A., Sutskever I., and Hinton G. E. Imagenet classification with deep convolutional neural networks. In \textit{Proc. NIPS}, pages 1097-1105, 2012.

[8] Zintgraf L. M., Cohen T. S., Adel T., and Welling, M. Visualizing deep neural network decisions: Prediction difference analysis. In \textit{Proc. ICLR}, 2017.

[9] Springenberg J. T., Dosovitskiy A., Brox T., and Riedmiller M. (2014). Striving for simplicity: The all convolutional net. \textit{arXiv preprint arXiv:1412.6806 }, 2014.

[10] Mahendran A., and Vedaldi A. Salient deconvolutional networks. In \textit{Proc. ECCV}, pages 120-135, 2016.

\end{document}